\documentclass{article}
 \usepackage{graphicx}
\usepackage{amsthm}
\usepackage{amsmath}
\usepackage{amssymb}
\usepackage{hyperref}

\title{Mathematics, word problems, common sense, and artificial intelligence}

\author{
Ernest Davis \\
Dept. of Computer Science \\ New York University \\ New York, NY 10012 \\
{\small davise@cs.nyu.edu}}

\begin{document}
\maketitle
\begin{abstract}
The paper discusses the capacities and limitations of current artificial
intelligence (AI) technology to solve word problems that combine 
elementary knowledge with commonsense reasoning. No existing AI systems
can solve these reliably. We review three approaches that have been 
developed, using 
AI natural language technology: outputting the answer directly, outputting
a computer program that solves the problem, and outputting a formalized
representation that can be input to an automated theorem verifier. We review
some benchmarks that have been developed to evaluate these systems and some
experimental studies. We discuss the limitations of the existing technology 
at solving these kinds of problems.
We argue that it is not clear whether these kinds of
limitations will be important in developing AI technology for pure mathematical
research, but that they will be important in applications of mathematics, and
may well be important in developing programs capable of reading and understanding
mathematical content written by humans.
\end{abstract}

\section{Mathematical problems and world knowledge}
\label{secIntro}

A central aspect of understanding mathematics 
is understanding how math
applies in real-world situations and how mathematical analysis can be
used to answer questions about real things. 
Consider, for instance, the following two elementary problems:

\begin{quote}
{\bf Problem 1:}
George has seven pennies, a dime, and three quarters. Harriet has four pennies
and four quarters. First, George gives Harriet thirty-one 
cents in exact change; then
Harriet gives him back exactly half of her pennies. How much money does
George now have?

{\bf Problem 2:}
You have an empty cylindrical open container whose inside has a diameter of 
8 centimeters and a height of 20 centimeters. and a pitcher with 200 ccs of water.
You first empty the pitcher into the cylinder, then
put a solid rock cube, 4 cm on a side,  into the container so that it is
sitting flush against the
bottom of the container. What is the height of the water in the container?
\end{quote}

Answering these questions requires knowledge of three kinds. First, and most
obviously, there is the elementary math involved: elementary 
arithmetic and solid geometry, respectively.
Second, one has to know the language; if these questions, as
printed above in English, are presented to a speaker who knows only Polish,
then they will not be able to answer, no matter how well they know the math.
Third, and most easily overlooked, 
each of these requires basic knowledge about the world,
and an understanding  of how the world is characterized mathematically. In
problem 1, you have to know the value of U.S. coins, but more fundamentally,
you have to understand the dynamics of ownership
and giving: If $A$ gives $X$ to $B$, then the result is that $B$ now owns $X$.
In problem 2, you have to understand the basic physics of liquids
and solids: solids maintain their shape over time,
liquids maintain their volume but not their shape, solids
and liquids do not overlap and so on.

In many simple problems, a significant part of the real-world understanding 
required is purely
``common sense'' --- the basic understanding of the realities of daily human
existence that is shared by all people past early childhood.\footnote{This
characterization of common sense has obvious significant problems and limitations,
but it will suffice for our discussion here. An in-depth discussion will be
found in \cite{DavisCSBench}.}  
When people work on problems of this kind, the commonsense knowledge
and reasoning involved is often so obvious that it goes entirely
unnoticed; what is difficult is finding the mapping to mathematics and
carrying out the mathematics. 

Other forms of world knowledge also enter into mathematical problem solving.
{\em Common knowledge\/}, such as the value of U.S. coins in problem 1, can
be culturally dependent and is sometimes
taught explicitly in schools, but is universal
among adults in the society under discussion. {\em Encyclopedic knowledge} is
specific facts about particular entities, such as the knowledge needed to
answer the question, ``How much older, in days, was George Washington than
Abraham Lincoln?'' {\em Expert knowledge\/} is knowledge held by experts but
not by lay people. Obviously these categories are vague and there
is no point in trying to delimit them with any precision.

A problem that calls on the combination of mathematics with world knowledge
will generally involve other cognitive capacities as well. 
If it is posed in language, then it requires comprehension of the 
language.  If the problem statement includes mathematical notation  or other
technical notation (musical, chemical, etc.).
then that notation must be understood. If an embodied agent 
must solve a problem in
the physical world, then understanding it requires perception, most 
commonly vision.
If the problem involves a diagram or graph, then it requires vision
plus an understanding of the graphical conventions. If an embodied agent
must carry out a task in the real world, then that requires perception
and manipulation.

It will be convenient to distinguish a number of categories of math problems.
A {\em symbolic mathematics problem\/} is one posed in mathematical notation with
minimal use of natural language, e.g. ``Solve $x^{3}-6x^{2}+11x-6=0$''.
A {\em word problem} is a mathematical problem with more than minimal use of
natural language.
A {\em purely mathematical word problem} is a word problem that 
makes minimal reference to any 
non-mathematical concepts, e.g. ``Find a prime number $p$ such that $p+36$ is
a square number."``What is the volume of a regular icosahedron with diameter 5?''
A {\em real-world word problem\/} 
is a word problem whose solution requires the use
of non-mathematical knowledge of some kind. 
A {\em commonsense word problem\/} 
(CSW) is a word problem that 
involves significant use of commonsense knowledge and perhaps also 
common knowledge, but not encyclopedic or expert knowledge; this is the most
important category for our purposes. Finally, an {\em elementary commonsense
word problem} is a CSW that requires only elementary math, however one
chooses to draw that line (generally elementary school math or high school math).
Again, obviously, the lines delimiting these categories are vague.

Real-world word problems generally and commonsense word problems in particular
are common in grade school math classes, rarer in introductory
college courses, and extremely rare in advanced college math courses.
They remain common in advanced courses
in other topics that use mathematics for applications; there, generally, the
knowledge involved is primarily expert knowledge, though commonsense
knowledge often forms an underlying and largely unobserved foundation.

\section{Artificial intelligence for word problems}
\label{secAI}

In recent years, artificial intelligence (AI) has made extraordinary progress 
in a wide range of domains, such as game playing, image tagging, speech
transcription, and machine translation. In particular, in recent years AI 
has achieved extraordinary success at certain kinds of language tasks, and, 
as discussed throughout this volume, noteworthy successes at a range of
mathematical tasks.  

One might reasonably suppose, therefore, that problems
as simple as problems 1 and 2 above  must be well within the capacities of
the current AI technology. However, that is not at all the case. {\em As of
the end of 2022, there does not exist any AI that can reliably solve
elementary CSWs.} The goal of this article is to elaborate on
that observation; to describe at a very high level how the current
technology works and what it can and cannot do,
and to discuss the challenges that lie ahead.
That is, this paper addresses the intersection of four 
issues: The use
of artificial intelligence (i) to solve problems requiring math (ii) and
commonsense reasoning (iii) posed in language (iv).

The challenge of developing AI systems that solve mathematical word problems
has been studied since Daniel Brobrow's STUDENT system of 1964 \cite{Bobrow}.
Skipping over the history of the problem over the subsequent 50 years, 
we will focus here on recent systems that apply AI technology to math
word problems.

Progress in AI over the last twenty years has been almost entirely driven by 
{\em corpus-based machine learning} (ML). 
Broadly speaking, in corpus-based machine learning,
a general-purpose computational architecture,
with little or no built-in knowledge of the domain or the task, is trained on
a corpus of relevant data, and it finds underlying patterns in the data that
it can use\footnote{I am including supervised learning, unsupervised learning.
and reinforcement learning in this general category.}
to carry out the task with some degree of accuracy. Almost all
current AI programs are built around 
a core of modules created using corpus-based
learning, often combined with additional, hand-crafted modules for particular
aspects of the task.

\subsection{Large Language Models}
\label{secMethods}

As discussed above, solving CWPs requires combining commonsense reasoning,
language abilities, and mathematics. For the last five or six years,
both natural language processing (NLP) and (more surprisingly) 
commonsense reasoning in
AI has been entirely dominated by one particular technology known as
large language models (LLMs) (in current AI parlance ``model'' means 
``computer program.''). Almost all recent work on math world problems 
has an LLM at the core.\footnote{An exception is Wolfram Alpha 
\url{https://www.wolframalpha.com/}, 
a manually constructed system with a large body of
mathematical algorithms and encyclopedic and expert knowledge 
from many scientific and sociological. It takes input posed in natural
language; however, its natural language abilities are comparatively limited
and it has little commonsense reasoning ability.}

Large language models are the most recent development of thirty years of 
study of how neural network AI technology (now often called ``deep 
learning'') can be applied to language-oriented tasks. We will not attempt
a technical description of their workings here (see \cite{transformer}), 
but we will
describe the characteristics most important for our discussion here.

The task that an LLM carries out is text generation in response to a prompt.
It generates its response word by word. That is: the user provides the AI
with an input text viz. a string of tokens (words, punctuation marks, and
special symbols) $w_{1} \ldots w_{n}$. The AI computes the most probable (or,
depending on the settings, a reasonably probable) next word $w_{n+1}$. It then
computes word $w_{n+2}$ that is most likely to follow $w_{1} \ldots w_{n+1}$.
And so on; it repeatedly computes the most probable next word, until it reaches
a halting state. (Chatbots such as ChatGPT 
have some additional mechanisms to
produce reasonable interactions.) This is known as the ``language modeling''
task.

The generation of words is thus driven by the conditional probability
of producing $w_{n}$ following $w_{1} \ldots w_{n-1}$,
$P(w_{n} \: | \: w_{1} \dots w_{n-1})$. The AI computes this in terms of 
a function $f_{\vec{\alpha}}(w_{1} \ldots w_{n})$ which is implicitly
encoded in the network structure of the AI.
This function $f_{\vec{\alpha}}$ is controlled 
by a parameter $\vec{\alpha}$, a real-valued vector.
In current systems, the dimension of $\vec{\alpha}$ 
ranges between about $10^{9}$ and $10^{12}$

In current system, during generation, the system only keeps track of the most
recent $k$ words where $k$ is several hundreds or thousands. These systems therefore
cannot read a long text of tens or hundreds of thousands of words and remember
those as a whole.

The training set for the LLM is a vast corpus of text 
($10^{10}-10^{13}$ tokens) downloaded from the Web. Detailed accounts of
the content of these corpora have not, in general, been published; but
it is safe to say that the majority consists of English language documents of
various kinds, but that the training copora also include substantial quantities
of texts in other languages, of software written in popular computer languages,
of mathematics in mathematical notation, of images, and other kinds of data.

Training the LLM consists in finding the value of the parameter $\vec{\alpha}$
such that the probability function 
$P(w_{n} \: | \: w_{1} \dots w_{n-1}) = f_{\vec{\alpha}}(w_{1} \ldots w_{n})$ 
matches the training set $T$ as closely as possible. That is, there is
an error function $E_{T}(\vec{\alpha})$ that measures the discrepancy between 
the system's predictions and the actual sequences in $T$ when the AI uses
$f_{\vec{\alpha}}$ to generate predictions. The value of $\vec{\alpha}$ 
is chosen to
(approximately) minimize $E_{T}(\vec{\alpha})$.
This minimization uses a gradient descent algorithm. The training
process requires enormous amounts of computation --- months of computation time
on large networks of computers --- as well as a substantial amount of 
expert human labor.
It is feasible only for large AI labs, not for individual
academic researchers or small companies.

This training procedure for an LLM is carried out only once, when the LLM is
created. However, it is possible to further ``fine-tune'' an LLM to a particular
task by further training on texts specifically relevant to that task.

The underlying architecture of the LLM, which determines the function 
$f_{\vec{\alpha}}$, and the training procedure are both very general in structure
and are built to carry out prediction for essentially any kind of input 
string with some kinds of regularities. They do not reflect {\em any\/}
knowledge of the characteristics of natural language generally, of any
particular language, of the external world that language describes, of the various
uses of language, or of any task other than string prediction. {\em All\/} 
the knowledge
of language, its meaning, and its content that the system possesses is in terms
of how best to carry out prediction over the training set.

As with all AI systems based on neural networks/deep learning, the function
$f_{\vec{\alpha}}$ is {\em opaque}, in the sense that it is extremely difficult
to find a relation between the vector $\vec{\alpha}$ 
and the behavior of the system,
either on a particular example or in general. It is therefore generally impossible 
to incorporate any of what is known about language, the world, or mathematics into
the system other than through training. It is also impossible to debug errors
as is done in ``conventional'' computer programming. All that can be done is
to retrain the system from scratch (or, in rare cases, to add a hand-crafted
patch at the back-end to deal with the problem.)

In earlier NLP research, different tasks --- question answering, chatbots,
summarization, 
information extraction, translation and so on --- were each handled separately
by systems built specifically for that purpose. However, in the last 
three or four years,
it has turned out that language modeling tasks can serve as a basis for all
of these; indeed for any task where the input and output are in language
\cite{FoundationModels}. For
some tasks, like chatbots, LLMs are currently the most effective technology.
For others, like translation, they are not competitive with specialized
technology but they are remarkably good.

It has also turned out, quite surprisingly, that the 
quality of answers generated by
the very large LLMs can often be significantly improved by including
a handful of examples of the kind of output desired in the prompt, or even by
including general directives such as ``Let's think about this step by step''.
Thus, if you want it to translate from German to English, it helps to include
a few examples of translation in the prompt;
if you want it to answer mathematical questions of a particular kind, it helps
to include a few examples of that kind of problem in the prompt. This is known
as ``few-shot prompting''; it was
first observed in the LLM GPT-3, released in 2020 \cite{GPT3}. It has given
rise to a new area of study: ``prompt engineering'', the creation of prompts
that guide LLMs to correct outputs of the proper form.

A final general remark is that LLMs are prone to so-called {\em hallucinations}: 
since they have no sense of underlying reality, they often generate text that 
reads smoothly but is actually incoherent. Depending on circumstances, a 
more savvy human reader can find this amusing, frightening, or misleading. 
Table~\ref{tabHallucination} shows a mathematical example generated by the 
LLM ChatGPT (experiment by the author, 1/23/2023).

\begin{table}
{\bf User query:} Please sketch the proof of Fermat's little theorem.

\vspace*{6pt}

{\bf ChatGPT output:} Fermat's Little Theorem states that if \verb|a| is an integer 
and \verb|p| is a prime, then the following equation holds:

\vspace*{6pt}

\verb|a^p = a (mod p)|

\vspace*{6pt}
This means that if you raise \verb|a| to the power of \verb|p| and divide the result by \verb|p|, 
the remainder will always be equal to \verb|a|.

\vspace*{6pt}
One way to prove Fermat's Little Theorem is by using mathematical induction.

\vspace*{6pt}
First, we prove the base case, where \verb|a = 1|. In this case, the equation holds 
true because \verb|1^p = 1|, and 1 divided by \verb|p| will always have a remainder of \verb|1|.

\vspace*{6pt}
Next, we assume that the equation holds true for some arbitrary value of \verb|a|, and we will prove that it holds true for \verb|a+1|. Using the assumption that the equation holds true for \verb|a|, we know that:

\vspace*{6pt}
\verb|a^p = a (mod p)|

\vspace*{6pt}
We can now multiply both sides of this equation by \verb|a|:

\vspace*{6pt}
\verb|a^(p+1) = a^2 (mod p)|

\vspace*{6pt}
Now, we can use the fact that \verb|a^2| can be written as 
\verb|(a+1)(a-1) +1|.

\vspace*{6pt}
So,

\vspace*{6pt}
\verb|a^(p+1) = (a+1)(a-1) +1 (mod p)|

\vspace*{6pt}
Since we know that \verb|a-1| is divisible by \verb|p|, and that \verb|p| is 
prime, we can say 
that \verb|(a+1)(a-1)| is also divisible by \verb|p|.
\vspace*{6pt}

Therefore,

\vspace*{6pt}
\verb|a^(p+1) = 1 (mod p)|
\vspace*{6pt}

which is what we wanted to prove.

\vspace*{6pt}
By mathematical induction, we have proved that Fermat's Little Theorem holds 
true for all integers \verb|a| where \verb|p| is a prime.

\vspace*{6pt}
It's important to note that Fermat's Little theorem is a fundamental theorem 
in number theory and it has many useful applications in cryptography, coding 
theory and some other fields. 
\caption{Hallucination from ChatGPT} 
\label{tabHallucination} 
\end{table} 

\% clearpage

\subsection{LLMs' abilities at common sense, math, and language}
Outside a mathematical context, LLMs achieve significant levels of success at
commonsense reasoning, though they are certainly not reliable, and they
have been steadily improving over time.
Unquestionably
LLM-based AIs are currently the most powerful and general, publicly 
available\footnote{The CYC project \cite{CYC}. begun in 1985 and still
ongoing, aims at encoding commonsense knowledge in a quasi-logical 
symbolic system and using automated inference to do commonsense 
reasoning. It is certainly large: it contains some millions of 
``concepts'' and tens of millions of propositions, and it
represents several hundred person-years of expert human labor. However, 
the contents are proprietary: no systematic description, evaluation, 
or benchmark test has ever 
been published, and the scientific literature on its workings is very small. 
Consequently, it is impossible for outsiders to know what its 
capabilities are. } 
technology for commonsense reasoning that have been built.

LLMs also have displayed some mathematical abilities. 
GPT-3, when originally released in 2020 \cite{GPT3},  was tested on simple
integer arithmetic problems. When prompted with a few correct examples, it 
achieved 100\% accuracy on two-digit addition (e.g. ``What is 35 plus 72?'') with
gradually diminishing accuracy as the number of digits increases, though only
21.3\% accuracy on problems involving two arithmetic operations and one-digit
numbers (e.g. ``What is (2 + 4) * 6?''). More recent systems have achieved 
significantly higher levels of success \cite{LLMArithmetic} 
but the general pattern remains. 

LLMs also can often, though unreliably, generate correct mathematics at a higher
level. The errors in the ``proof'' of Fermat's little theorem
table~\ref{tabHallucination} 
are amusing, but one should note what ChatGPT got right here.
The theorem was correctly stated. The algebraic manipulations are correct,
on the unjustified assumption that a-1 is divisible by p. The general structure
of an inductive proof --- prove the base case a=1, prove that if the theorem
holds for a then it proves for a+1 --- is followed, though ChatGPT fails
to notice that the statement that it ``proved'' was not the statement it wanted
to prove.

Finally, LLMs are fundamentally language technology, and in many respects their
purely linguistic 
abilities are extraordinary and flawless, though not generally very 
creative. The current generation of
LLMs can produce essays several paragraphs long that, in their language use,
are often indistinguishable from capable human writers. I have never seen them
make a grammatical error or malapropism in writing modern 
English.\footnote{I did once catch ChatGPT using ``doth'' as a plural 
verb in trying
to write a Shakespearean pastiche.} 

LLMs thus have at least some facility at the three components of CSW:
common sense, math, and language. However, putting these together is a different
matter.

\subsection{Using an LLM to solve a word problem}

There are three major approaches to using an LLM to solve mathematical
word problems. They all take the problem, possibly supplemented with a
prompt as input to the LLM; they differ in the nature of the output.  

The most straightforward approach is simply to ask the LLM to produce the
answer to the problem. This can work well with simple mathematics, though
it breaks down when more complicated math is required.

\subsubsection{Translating a word problem into executable code}

A second approach is to have the LLM produce computer code that can then
be executed to compute the answer. Generating code is, in fact, the
most successful practical applications of LLMs to date. AI programs like
OpenAI's Codex \cite{Codex} and GitHub CoPilot \cite{CoPilot} are 
used by professional programmers as assistants to help write code;
some experts have enthusiastically reported that they save significant
time and effort. These AIs are particularly effective at generating
code of standard but non-obvious form, such as 
finding the names and arguments to library functions based on a verbal
description;  they thus save the programmer a tedious search through 
documentation. They are much less reliable in generating programs of 
significant length of complex structure. 

Drori et al. \cite{Drori} used Codex to generate Python code that would compute
the answers to a collection of problems drawn from undergraduate math
courses at MIT and Columbia. Their system involved the following steps.
First, the problem was modified by a hand-crafted automated front end into a
more standardized natural language form; generally this required only adding
a few stock phrases such as ``Use sympy".
Second, the modified problem was then input to Codex, which output Python code.
Finally, if the Python code did not give the correct answer, the system searched
for similar examples, with solutions, in the training corpus. These were
used to create a few-shot prompt, and the problem was attempted again with
the new prompt.

Drori et al.  claimed a success rate of 81\%; however,
for a number of reasons, that figure is highly misleading \cite{DavisOnDrori}.
Two successful examples, chosen as illustrations in \cite{Drori}, 
are shown in table~\ref{tabDrori}.

\begin{table}
{\bf Question:} Find the derivative of the function using the definition
of a derivative. \\
\verb|f(x) =Y (x**2-1)/(2*x-3)|.

\vspace{6pt}

{\bf Input:} Using sympy find the derivative of the function using the definition
of a derivative. \\
\verb|f(x) =Y (x**2-1)/(2*x-3)|.

\vspace{6pt}

{\bf Program:}
\begin{verbatim}
import smpy as sp

x = sp.Symbols('x')

f = (x**2-1)/(2*x-3)

print(sp.diff(f,x))
\end{verbatim}

{\bf Answer:} 2*x/(2*x-3)-2*(x**2-1)/(2*x-3)**2

\vspace{14pt}
{\bf Problem:} One generates a number x from a uniform distribution on
the interval [0,$\theta$]. One decides to test H0: $\theta=2$ against HA:
$\theta \neq 2$ by rejecting H0 if x$\leq$0.1or x$\geq$1.9. Compute
the probability of a type 1 error.

\vspace{6pt}

{\bf Input:} One generates a number x from a uniform distribution on
the interval [0,$\theta$]. One decides to test H0: $\theta=2$ against HA:
$\theta \neq 2$ by rejecting H0 if x$\leq$0.1or x$\geq$1.9. Using simulation
compute the probability of a type 1 error.

\vspace{6pt}

{\bf Program}
\begin{verbatim}
import numpy as np

def generate_x(theta)
    return np.random.uniform(0,theta)

def test_hypothesis(x)
    if x <= 0.1 or x >= 1.9:
        return TRUE
    else:
        return FALSE


def simulate_type_1_error(theta,n)
    count = 0
    for i in range(n)
        x = generate_x(theta)
        if test_hypothesis(x):
            count += 1
    return count / n

print(simulate_type_1_error(2,100000))
\end{verbatim}

\vspace{6pt}
{\bf Answer:} 0.1

\caption{Translating word problems into Python code. From \cite{Drori}.}
\label{tabDrori}
\end{table}

A few points about these examples should be noted. In both examples, 
the system ended up altering the original specification: in the first,
it did not use the definition of the derivative; in the second, presumably
the author of the problem did not intend that simulation should be used.
In the first example, Codex is carrying out an almost trivial transformation
from the language expression to a hand-constructed sympy library function. 
In the second
example, the transformation from language to code is not at all trivial, but
it is safe to say that Codex is here closely imitating some code that was in its 
training set.

An inherent limitation of this approach is that it carries out the translation
in strict order: It first generates the code, then executes it. Presumably all
the numerical calculations are done in the execution part. If so, then the 
code must work for all numerical values of the parameters. This can end up turning
an easy problem into a well-nigh impossible one (or even an absolutely 
impossible one). Many commonsense problems are easy for some
of the numerical parameters involved but extremely difficult 
for other values. If the problem
is stated with values for which the problem is easy, then generating the code
for the difficult cases is pointless, but may well be 
impossible to avoid in this
approach. As one example of many, consider the problem of determining whether 
an object of complex shape fits through a cylindrical hole. If the object is
small enough (e.g. its diameter is less than the diameter of the cross section
of the hole) , then trivially, it can fit through. If it is too
large (e.g. it contains a sphere whose diameter is larger than the diameter
of the cross section of the hole) then, trivially, it cannot fit through.
Otherwise the calculation is more difficult, and for tight fits, it can be 
{\em much\/} more complicated. Certainly the output code can be written so that,
in simple cases, the code for difficult cases is not {\em executed\/}, but the
point is, if the problem at hand is an easy case, it should not be necessary to
even {\em generate\/} the code.

\subsubsection{Translating word problems into logical specifications}
\label{secCode}

A third approach to solving word problems is built around 
{\em autoformalization\/}, the process of automatically translating
mathematical specifications given in natural language into a formal
specification that then can be given to a formal verification
system. Wu et al \cite{autoformalization}  used two LLMs, PaLM and Codex,
to translate word problems into the formal notation used by the
interactive proof assistant Isabelle. Since the training set for the LLMs
presumably contains only a comparatively small quantity of Isabelle, 
they provided long few-shot prompts (different prompts for different
areas of mathematics) with ten examples.
They tested their systems on examples taken from
benchmark collections MATH and MiniF2F, both of which contain mathematical
competition problems.  They found that the system created a perfect translation
25\% of the time. Problems that require finding a specific value as an answer
were reworded by adding, ``The final answer is $A$'' where $A$ is the correct
answer at the end of the natural language formulation.
(This is reasonable for testing the adequacy 
of the translation of $\alpha$ into a formal notation; it would not, of course,
be reasonable for testing the capacity of an AI to {\em solve\/} the
problem.)
Table~\ref{tabAutoForm} shows two examples of two problems translated
into Isabelle. Note that the second example is in fact incorrect, though it
is marked in \cite{autoformalization}
(p. 23) as ``correct''. 

\begin{table}
{\bf Question} Natural language version: 
``When all the girls at Madeline's school line up in rows of
eight, there are seven left over. If instead they line up in rows of four,
how many are left over? The final answer is 3.'' Translate the natural 
language version to an Isabelle version

\vspace{6pt}

{\bf PaLM output:}
\begin{verbatim}
theorem
  fixes n::nat
  assumes "n mod 8 = 7"
  shows "n mod 4 = 3"
\end{verbatim}

\vspace{14pt}
{\bf Question} Natural language version: 
"A map of the town that Annie, Barbara, and Charlie live in can
be represented by the Cartesian plane. Annie is located at (6, -20)  and 
Barbara is located at (1, 14). They agree to meet at the closest point 
that is equidistant from their current locations and walk upwards together 
to get to Charlie’s location at ($\frac{7}{2}$ , 2). How many units
upward do Annie and Barbara walk together to get to Charlie? 
The final answer is 5."
Translate the natural language version to an Isabelle version:

\vspace{6pt}
{\bf Codex Output:}
\begin{verbatim}
theorem
  fixes a b c :: "real \<times> real"
  assumes h0 : "a = (6, -20)
    and h1 : "b = (1, 14)"
    and h2 : "c = (7 / 2, 2)"
  shows "dist a c + dist b c = 5"
\end{verbatim}
\caption{Examples of autoformalization from \cite{autoformalization}.
The translation in the second example is incorrect.}
\label{tabAutoForm}
\end{table}


\subsection{Benchmarks}
\label{secBenchmarks}
Standard collection of elementary math problems serve as benchmarks
to compare the capacities of different AI systems with one another
and with humans, to measure progress in the technology, and to serve  
as targets for research.

Numerous benchmark collections of word problems have been 
assembled \cite{LILA}. The mathematical difficulty ranges from 
elementary school to International Mathematical Olympiad and introductory
college math courses. The language use ranges from non-linguistic to 
moderately complex. Some require no world knowledge; others requires some
combination of commonsense, common, expert, and encyclopedic knowledge. 
Question formats include open form question answering, 
fill-in-the-blank, multiple choice, and proof construction. 

One high-quality collection of elementary CSPs
is SVAMP (Simple Variations on Arithmetic Math Problem) dataset 
\cite{SVAMP}.
The problems in SVAMP involve only commonsense and common knowledge; the
language is simple; the math required is either one or two arithmetic 
operations.
The numbers involved are positive integers between 1 and 999. 
Each example is given in a number of paraphrases
and variants, in order to avoid the kinds of artifacts discussed below in
section~\ref{secArtifacts}. (AIs presented
with multiple examples do not, ordinarily, carry information from one
to the next; each is considered from scratch.\footnote{Chatbots such as ChatGPT
are exceptions to this general rule.})
Thus, the fact that there are multiple similar problems given in sequence does
not affect behavior as it would with human experimental subjects.)
Table~\ref{tabSVAMP} shows some example problems in
SVAMP. There are 1000 such problems in SVAMP.

\begin{table}
Allan brought two balloons and Jake brought four balloons to the park. How many balloons
did Allan and Jake have in the park?

\vspace*{6pt}

Allan brought two balloons and Jake brought four balloons to the park. How many more
balloons did Jake have than Allan in the park?

\vspace*{6pt}

Jack had 142 pencils. Jack gave 31 pencils to Dorothy. How many pencils does Jack have now?

\vspace*{6pt}

Dorothy had 142 pencils. Jack gave 31 pencils to Dorothy. How many pencils does Dorothy
have now?
\vspace*{6pt}

 Jack had 142 pencils. Dorothy had 50 pencils. Jack gave 31 pencils to Dorothy. How many
pencils does Jack have now?
\caption{Examples from SVAMP dataset}
\label{tabSVAMP}
\end{table}


Another notable benchmark for math word problems, named {\sc L\=ila}, 
is a large, structured collection, covering a range
of linguistic complexity, mathematical difficulty, and requisite background
knowledge \cite{LILA}
The collection was assembled
by combining twenty-three previously existing benchmarks collections and 
standardizing their formats. Table~\ref{tabLILA} shows the different 
categories of math problems included, with examples.

\begin{table}
\begin{tabular}{l|l|l} \\ \hline
& {\bf Category} & {\bf Example} \\ \hline

1. & {\small Basic math: addition, subtraction,} & 
{\bf Question}
{\small If Jimbo is 484 feet away from a beetle and quarter} \\
& {\small fact-based QA etc.} & 
{\small of 827 feet from a grasshopper, which insect will seem bigger to him? }
\\
& & {\small "Option 1": beetle, "Option 2": grasshopper. 
{\bf Answer:} Option 2} \\ \hline

2. & {\small Muldiv: multiplication, division along} &
{\small {\bf Question:} Mrs. Hilt bought 2 pizzas.  Each pizza had 8 slices.}
\\ 
& {\small with addition, subtraction etc.} & 
{\small So, she had \rule{0.5cm}{0.5pt} total slices of pizza.} 
{\bf Answer:} 16  \\ \hline

3. & {\small Number theory: prime, power, negation,} &
{\small {\bf Question:} How many numbers are divisible by both 2 and 3 
up to 300?} \\ 
& {\small modulus and other operators etc.} & 
{\small {\bf Answer:} 50} \\ \hline

4. & {\small Algebra: equations, functions,} & 
{\small {\bf Question:} 
The sum of the three smallest of four consecutive integers is }
\\
& {\small polynomials, series etc.} &
{\small 30 more than the largest integer. What are the four} \\
& & {\small consecutive integers ? {\bf Answer:}
15.0 }\\ \hline 
5. & {\small Geometry: triangles, polygons,} &
{\small {\bf Question:} 
A hall is 6 meters long and 6 meters wide. If the sum of the } \\
& {\small 3D structures etc.} & 
{\small areas of the floor and the ceiling is equal to the sum of the 
areas of four } \\
& & {\small walls, what is the volume of the hall (in cubic meters)? 
{\bf Answer:} 108} \\ \hline

6. & {\small Statistics: binomial, divergence, mean,} &
{\small {\bf Question:} There are 11 boys and 10 girls in a class. 
If three students are} \\
& {\small median, mode, variance etc.} &
{\small selected at random, in how many ways that 
3 girl and 2 boys are selected? [sic]
} \\
& & {\small {\bf Answer:} 6600} \\ \hline

7. & {\small Calculus: differentiation, integration,} &
{\small {\bf Question:} Let g(y) = 9*y**4 + 25*y**2 + 6.}  \\
& {\small gradient, series expansion etc} &
{\small Let s(d) = 1 - d**4. Let x(t) = -g(t) + 6*s(t). }
\\
& & {\small 
What is the third derivative of x(f) wrt f? {\bf Answer:} -360*f }
\\ \hline

8. & {\small Linear algebra: vectors, dot products,} &
{\small {\bf Question:} Problem: Convert the following matrix to reduced 
row echelon form }
\\
& {\small Eigen vectors, matrices etc.} & 
\includegraphics[width=3in]{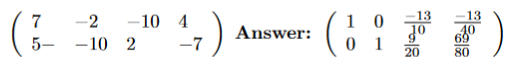} \\ \hline

9. & {\small Advanced math: heuristics required}  &
{\small {\bf Question:} Let $f(x)=2^{x}$. Find $\sqrt{f(f(f(f(1))))}$. 
{\bf Answer:} 256.} \\
& {\small along with probability, statistics, or}  & \\
& {\small algebra, Olympiad level problems} & \\ \hline
\end{tabular}
\caption{Categories of math difficulty in {\sc L\=ila}, together with examples
\cite{LILA}}
\label{tabLILA}
\end{table}


It is certainly disconcerting that three of the questions shown in
table~\ref{tabLILA} are flawed. The answer to question 1 is in fact
indeterminate; it can depend on the species of beetle
(beetles vary in length from a fraction of a millimeter to 20 cm), the angle
from which the insect is being viewed, whether the grasshopper has its wings
spread or shut, and whether perceived area or perceived
length is the measure in question. 
It is also not at all clear to me why the beetle/grasshopper problem
is considered ``basic arithmetic'' whereas the much simpler pizza problem is
put into the presumably more advanced category ``Muldiv''.
The answer ``15'' given for question 4 
does not follow the format specified in the question, ``What are the four
consecutive integers?'' As worded, Question 6 is both meaningless 
and ungrammatical; presumably 
``three'' is a typo for ``five''.
Regrettably, problems with flaws of these kinds,
and much worse, are all too common in AI benchmarks \cite{DavisCSBench}.
One cannot have much confidence in measurements made with such flawed
instruments.

\subsubsection{Untested abilities}
\label{secUntested}
Many of the benchmark collections used to test AI math abilities are taken 
from tests developed for human students --- math course problems, IQ tests,
SAT tests, Math Olympiad tests, and so on. However, these generally take
for granted that the test-takers have certain basic abilities; this is
a safe assumption for humans but not for AIs, whose abilities are very
different \cite{DavisOECD}. Other benchmarks have been developed specifically
for AIs, but these tend to focus on some abilities and omit others. Consequently,
many categories of problems can go untested.

For example, apparently none of the benchmark sets for mathematical ability
test an AI's ability to count. Surely, an AI system that can solve (some)
advanced math problems can count up to four?  Not necessarily, as determined
in informal test (table~\ref{tabCount}).  Likewise, many forms of 
commonsense mathematical inference that draw on a basic of time, space, and
physical reasoning are not tested in any existing AI benchmark.

\begin{table}
{[User]} What is the fourth word in the sentence, ``I am not what I am''?

\vspace{6pt}

[ChatGPT] The fourth word in the sentence, ``I am not what I am'' is ``not.''

\vspace{6pt}

[User] The children arrived one by one to the birthday party: First Tome, then
Mary, then Barbara, then Harry, then Ida. Who was the fourth child to arrive?

\vspace{6pt}

[ChatGPT] Barbara was the fourth child to arrive to the birthday party.

\vspace{6pt}

[User] How many words are in the sentence, ``I am not what I am"?

\vspace{6pt}

[ChatGPT] There are 7 words in the sentence ``I am not what I am."

\vspace{6pt}

[User:] Barbara, Mary, and Harry had red balloons. Ted had a blue balloon. 
Sam also had a red balloon. How many children had red balloons?

\vspace{6pt}

[ChatGPT:] Barbara, Mary, Harry, and Sam had red balloons. In total, there
were 5 children with red balloons.
\caption{Some tests carried out by the author on ChatGPT, December 30, 2022. 
ChatGPT did sometimes answer these kinds of questions, but it often answered
them incorrectly.}
\label{tabCount}
\end{table}


\subsection{State of the art with ML-based systems}
In experiments carried out in 2021, the creators of SVAMP found that, for
the three state-of-the-art AI programs they tested, the best performance
on problems with one arithmetic operation was 51.9\%; the best performance
on problems that required combining two arithmetic operations was 33.9\%. 
(These were achieved by different programs.)

A recent (October 2022) extensive study of the state of the art in LLM-based
models for math word problems is reported in \cite{LILA}, using the {\sc L\=ila}
benchmark described above. 
A new AI for math word problems, named {\sc Bh\=askara} was created, by
fine tuning GPT-Neo (a variant of GPT-3), trained on code as well as text) 
on a training set drawn from {\sc L\=ila}. More precisely,
two variants of {\sc Bh\=askara} were created. {\sc Bh\=askara-A} outputs the
answer directly; {\sc Bh\=askara-P} outputs Python code that computes
the answer.

Six AI programs were then tested on a test set\footnote{More precisely,
they were each tested on two different test sets drawn {\sc L\=ila}: 
one test set where the problems were drawn from the same original benchmark as  
those in the training set (the ``independent identically distributed (IID) state)
and a different test set where the problem of the same category were drawn
from a different original benchmark (the ``out of distribution'' OOD state).
Table~\ref{tabLilaResults} gives only the results for IID; these were generally,
but not always, higher than OOD.} drawn from {\sc L\=ila} and disjoint
from the training set: 
(i) GPT-3, with few-shot prompting;
(ii) Codex, with few-shot prompting;
(iii) Neo-A, i.e. GPT-Neo configured to output the answers;
(iv) Neo-P, i.e. GPT-Neo configured to output Python code;
(v) {\sc Bh\=askara-A}; (vi) {\sc Bh\=askara-P}. Table~\ref{tabLilaResults}
shows the best and worst results on each category of math problem, as well as the
results on problems considered ``commonsense''.

\begin{table}
\begin{center}
\begin{tabular}{l|l|l|l|l}
{\bf Category} & {\bf Best } & {\bf Best}  & {\bf Worst }
& {\bf Worst} \\
& {\bf result} &{\bf program}  & {\bf result} & {\bf program} \\ \hline
Basic math & 0.79 & Codex & 0.61 & Neo-P \\ \hline
Muldiv     & 0.69 & Codex & 0.16 & {\sc Bh\=askara-A} \\ \hline
Number theory & 0.47 & Codex  & 0.12 & {\sc Bh\=askara-A} \\ \hline
Algebra & 0.60 & Codex & 0.20 & {\sc Bh\=askara-A} \\ \hline
Geometry & 0.30 & {\sc Bh\=askara-A} & 0 & Codex  \\ \hline
Statistics & 0.95 & {\sc Bh\=askara-P} & 0.02 & {\sc Bh\=askara-A } \\ \hline
Calculus & 0.93 & Codex & 0.10 & {\sc Bh\=askara-A} \\ \hline
Linear algebra & 0.81 & Neo-P & 0.12 & GPT-3 \\ \hline
Advanced math & 0.47 & Codex & 0.02 & {\sc Bh\=askara-A} \\ \hline
Commonsense & 0.79 & GPT-3 & 0.62 & Neo-P \\ \hline
\end{tabular}
\end{center}
\caption{Results from tests on {\sc L\=ila}. From \cite{LILA}, table 3.
Results are for the ``IID'' regime. NEO-A, which never does well, is excluded
from the listing of worst results}
\label{tabLilaResults}
\end{table}

Conspicuous in table~\ref{tabLilaResults} is that both absolute performance and
relative performance between programs varies very much between categories. 
Each of the five systems is represented in the ``Best'' column; four of
them are also in the ``Worst'' column and {\sc Bh\=askara-P} misses that
by a hair, having scored 0.079 in the Geometry category. The wide
divergence between program-based systems and answer-based systems in 
the Statistics category is easily explained: these problems are mostly easy
if you can call library functions and very difficult otherwise.

\subsection{General caveat}
It cannot be assumed that, because an AI does well on a class of problems,
it can do well on a seemingly easier problem. As discussed in
section~\ref{secUntested}, the powerful AI ChatGPT, which does reasonably
well on various simple math tasks, cannot reliably count to four.
Likewise, it cannot be assumed that an AI that can carry out each of two
components of a given task separately; as we will discuss below, LLMs in general
do much worse on problems that involve two arithmetic operations than those
that require one, both in word problems and in purely mathematical problems.  
The capacities and incapacities of
these kinds of programs do not at all resemble those of human beings.

\subsection{Artifacts}
\label{secArtifacts}

A persistent problem in corpus-based ML, in all its applications, 
is that the patterns that the AI finds do not actually reflect the fundamental
characteristics of the problem, but rather superficial regularities in the
training data, known as ``artifacts''. For example, AIs trained over corpora
of medical images have ``learned'' to associate diagnoses with features of the 
different imaging devices used at different hospitals rather than the actual
content of the image. Artifacts are particularly likely to arise if a large
data corpus is partitioned into a training set, used to train the ML system,
and a test set, used to evaluate it; any superficial regularity in the corpus
as a whole can be learned from the training set and then applied to the 
test set. In general, avoiding artifacts in AI systems demands great care
in constructing corpora and carrying out tests.

These kinds of problems have, in fact, arisen in AI for math word problems.
For example, in one experiment \cite{SVAMP}, an AI trained on the 
benchmark ASDiv-A seems to have learned\footnote{AI systems that used
``deep learning'' and similar learning techniques are opaque; characterizing
what they have learned in terms of these kinds of rules is always 
approximate and relies on indirect evidence.}
that if the word ``every'' appears in a problem, it should
multiply the two numbers involved, whereas if the word ``each'' appears,
it should divide them.  The AI therefore got the right answer on questions
like
``John delivered 3 letters at every house. If he delivered for 8 houses, how many letters did John deliver?'' and 
``Sam made 8 dollars mowing lawns over the Summer. He charged 2 bucks for each lawn. How many lawns did he mow?'', but the wrong answer on questions like,
``John delivered 3 letters at every house. He delivered 24 letters in all. How many houses did John visit to deliver letters?'' and
``Sam mowed 4 lawns over the Summer. If he charged 2 bucks for each lawn, how much did he earn?.'' Apparently problems conforming to the rule were more common
in the dataset than those violating the rule. Thus, if an AI is trained on
a training set collected from this corpus, and then tested on a separate
group of problems from the same corpus, it may well be able to use these kinds
of regularities to achieve a fairly high success rate without at all 
understanding the actual meaning of the problem or its relation to the
mathematical operations.

\section{Further forms of common sense}
In addition to non-mathematical 
common sense about the world, there are various forms 
of commonsense knowledge
about mathematics itself and its relation to the world. To date, 
as far as I know, these have not been 
addressed in any AI systems. However, ultimately, in AIs that truly 
understand what mathematics is about they will have to be addressed.

One issue is the degree of precision that is reliable or meaningful in
characterizing a real-world situation. Specifying numbers to meaningless
degrees of precision, of course, is a common error
in human writers, but one may hope that AIs can do better. It happens to
be a pet peeve of mine.  One example that
come to hand: The accuracy numbers that I have given in 
table~\ref{tabLilaResults} are given to three-digit precision in 
\cite{LILA}. Now, three-digit precision is not even warranted statistically;
that would require a sample of size approximately one million. In 
table~\ref{tabLilaResults} I truncated these figures to two digits. However, 
as any
kind of replicable and meaningful measure of the quality of current
AI programs on math problems
of various categories, probably one digit is too many; my guess is that the
precision is more like a half a digit (two bits). Probably a program that scores
0.7 on a category is better than one that scores 0.2 and possibly it is 
better than one that scores 0.5; but my hunch is that
would be risky to claim more than that. (The wide range between IID tests and 
OOD tests for many systems on many categories and especially the fact
that OOD measures are sometimes substantially larger than IID values is
a further indication that these numbers are not very meaningful.)

A second, very difficult, issue is deciding, in trying to answer a question
about a real-world situation, how best to model it mathematically 
and how much detail to include.
Consider, for instance, a pendulum swinging on a string. The simplest model,
from a high school physics class,
is the harmonic oscillator $\ddot{\theta}(t) = -gl \theta(t)$. A more
sophisticated model changes this to  
$\ddot{\theta}(t) = -gl \sin(\theta(t))$.  But
under other circumstances, you might need to take into account: the
three dimensional angle and rotation around a vertical axis (if it is
swinging freely in 3D), the three
dimensional shape of the bob (if it is not a sphere), 
the physical characteristics of the string (e.g.
for a yo-yo, or if you plan to cut the string), the frictional forces (if you
are using it for a grandfather clock and need to supply energy), the rotation
of the earth and the latitude (for Foucault's pendulum), the presence of
other nearby massive objects (if, like Pierre Bouguer in 1735, you are camped
near the base of a mountain and using a plumb line to determine the vertical)
and so on \cite{Pendulum}.  The choice depends on
the physical circumstances, the information you have available, the kind of 
question you are asking, and the precision and certainty that you require in
the answer.

\section{Relevance to mathematical research}
As we have seen, current AI technology has serious limitations in its ability
to engage with even very simple real-world mathematical problems posed in
natural language. However, it is far from clear how relevant those limitations
to an AI's ability to do advanced mathematics. One might naturally 
suppose that
a technology that cannot reliably count up to four in a simple three-sentence
narrative is not about to have much
impact on groundbreaking mathematical research, but that is not a safe  
assumption. 

I doubt, for instance, that the limitations discussed here are
very relevant to the prospects of building AIs that can construct
formal proof of statements in pure mathematics given formal specifications.
Human mathematicians may well draw on their commonsense knowledge in 
understanding advanced mathematics and searching for proofs, though the
degree to which they do this is unknown. (It probably varies widely across
individuals.) But that does not at all imply that AIs would need to do likewise.
There are no AI programs that can reliably recognize and name concepts like
a ``fork'' or a piece being ``pinned'' in chess, or that can read and discuss
an article in a chess journal, but nonetheless they play very good chess.

However, there are important aspects of research mathematics where I expect
that
these limitations will be important. One is applications of all kinds. The
mathematization of real-world domains and tasks generally conceals the
commonsense reasoning involved and gives the illusion of a rigorously
formal process, but it does not actually eliminate the need for commonsense
understanding. This is true even in physics, at least on the experimental
side. Understanding how the LIGO gravitational wave detector works involves
a commonsense understanding of the components in addition to a lot of
technical understanding; you cannot prove the correctness of the experimental
device from first principles. In areas such as biology, medicine, and the
social sciences and cognitive sciences, commonsense reasoning is even
more unavoidable.

I would also conjecture, with much less confidence, that the limitations
discussed here would raise challenges to building an AI that can read and
understand (human-written) mathematical articles. I do not know of any
systematic analysis of the cognitive processes required to read an extended 
proof, but it seems likely that they draw significantly on basic commonsense
understanding. Again, no doubt it varies by field and by proof style. It seems
likely that proofs that draw strongly on one's visual sense and spatial, such as
(at the elementary level) those in \cite{Nelsen} or the beautiful proof
of Fermat's little theorem in \cite{Ellenberg} will be particularly difficult.
{\em Writing} mathematical proofs that are human readers find intelligible
and well-written
may well be easier than reading them; it is often found, in AI, that
generation is easier than comprehension. 

\subsection*{Acknowledgements}
Thanks to Scott Aaronson, Yuling Gu, and Doug Hofstadter for helpful feedback.

\end{document}